\documentclass[letterpaper, 10 pt, conference]{ieeeconf}  

\IEEEoverridecommandlockouts                        

\usepackage{cite}
\usepackage{amsmath,amssymb,amsfonts}
\usepackage{algorithmic}
\usepackage{graphicx}
\usepackage{multirow}
\graphicspath{{figures/}}
\usepackage{textcomp}
\usepackage{booktabs}
\usepackage{xcolor}
\usepackage{makecell}

\title{\LARGE \bf
EndoDDC: Learning Sparse to Dense Reconstruction for Endoscopic Robotic Navigation via Diffusion Depth Completion
}

\author{Yinheng Lin$^{1\dagger}$, Yiming Huang$^{1\dagger}$, Beilei Cui$^{1}$, Long Bai$^{1}$, Huxin Gao$^{1}$, Hongliang Ren$^{1}$ and Jiewen Lai$^{1*}$
\thanks{This work was supported in part by Guangdong Basic and Applied Basic Research Foundation (Grant No. 2025A1515011594); National Natural Science Foundation of China (Grant No. 62403402); NSFC Distinguished Young Scientists Fund – Category A (Grant No. T252500134); Hong Kong Research Grants Council (Grant Nos. C4026-21GF, R4020-22, 14200425, 14206125, 14204524, 14203323) \textit{(*Corresponding author: Jiewen Lai)}} 
\thanks{$^1$All authors are with the Department of Electronic Engineering, The Chinese University of Hong Kong, Hong Kong, China. \tt{\{linyinheng, yhuangdl, beileicui, b.long\}@link.cuhk.edu.hk}, \tt{\{hxgao, hlren, jwlai\}@ee.cuhk.edu.hk}}
\thanks{$^{\dagger}$These authors contributed equally to this work.}
}

\begin{document}

\maketitle
\thispagestyle{empty}
\pagestyle{empty}

\begin{abstract}


Accurate depth estimation plays a critical role in the navigation of endoscopic surgical robots, forming the foundation for 3D reconstruction and safe instrument guidance. Fine-tuning pretrained models heavily relies on endoscopic surgical datasets with precise depth annotations. While existing self-supervised depth estimation techniques eliminate the need for accurate depth annotations, their performance degrades in environments with weak textures and variable lighting, leading to sparse reconstruction with invalid depth estimation. Depth completion using sparse depth maps can mitigate these issues and improve accuracy. Despite the advances in depth completion techniques in general fields, their application in endoscopy remains limited. To overcome these limitations, we propose EndoDDC, an endoscopy depth completion method that integrates images, sparse depth information with depth gradient features, and optimizes depth maps through a diffusion model, addressing the issues of weak texture and light reflection in endoscopic environments. Extensive experiments on two publicly available endoscopy datasets show that our approach outperforms state-of-the-art models in both depth accuracy and robustness. This demonstrates the potential of our method to reduce visual errors in complex endoscopic environments. Our code will be released at \texttt{https://github.com/yinheng-lin/EndoDDC}.
\end{abstract}

\section{Introduction}
Accurate navigation of endoscopic surgical robots is fundamental to minimally invasive surgery, enabling surgeons to achieve improved precision, flexibility, and patient outcomes~\cite{ min2025innovating}. A critical challenge for navigation of these systems is accurate depth estimation. These robots acquire depth maps from 2D images and convert them into 3D representations, which is essential for improving spatial awareness, enabling autonomous path planning, and supporting real-time navigational decisions within complex anatomical structures~\cite{Intro_1.3_2Dto3D}.
Despite their promising potential, accurate depth perception remains a primary challenge for the navigation of endoscopic surgical robot systems. One common approach involves fine-tuning state-of-the-art (SOTA) models on surgical datasets \cite{Intro_3.1_finetune}. However, this requires dense depth ground truth that is difficult to acquire. The endoscopic environment, characterized by textureless tissue surfaces and specular reflections, fundamentally limits data acquisition. Consequently, depth sensors such as Time-of-Flight (ToF) \cite{Intro_2.1_ToF} and stereo endoscopes \cite{stereo_endoscopes} provide only sparse and incomplete depth measurements. While these sparse points are accurate, they are insufficient for the model fine-tuning to reconstruct a complete and dense 3D scene.


\begin{figure}
    \centering
    \includegraphics[width=\linewidth]{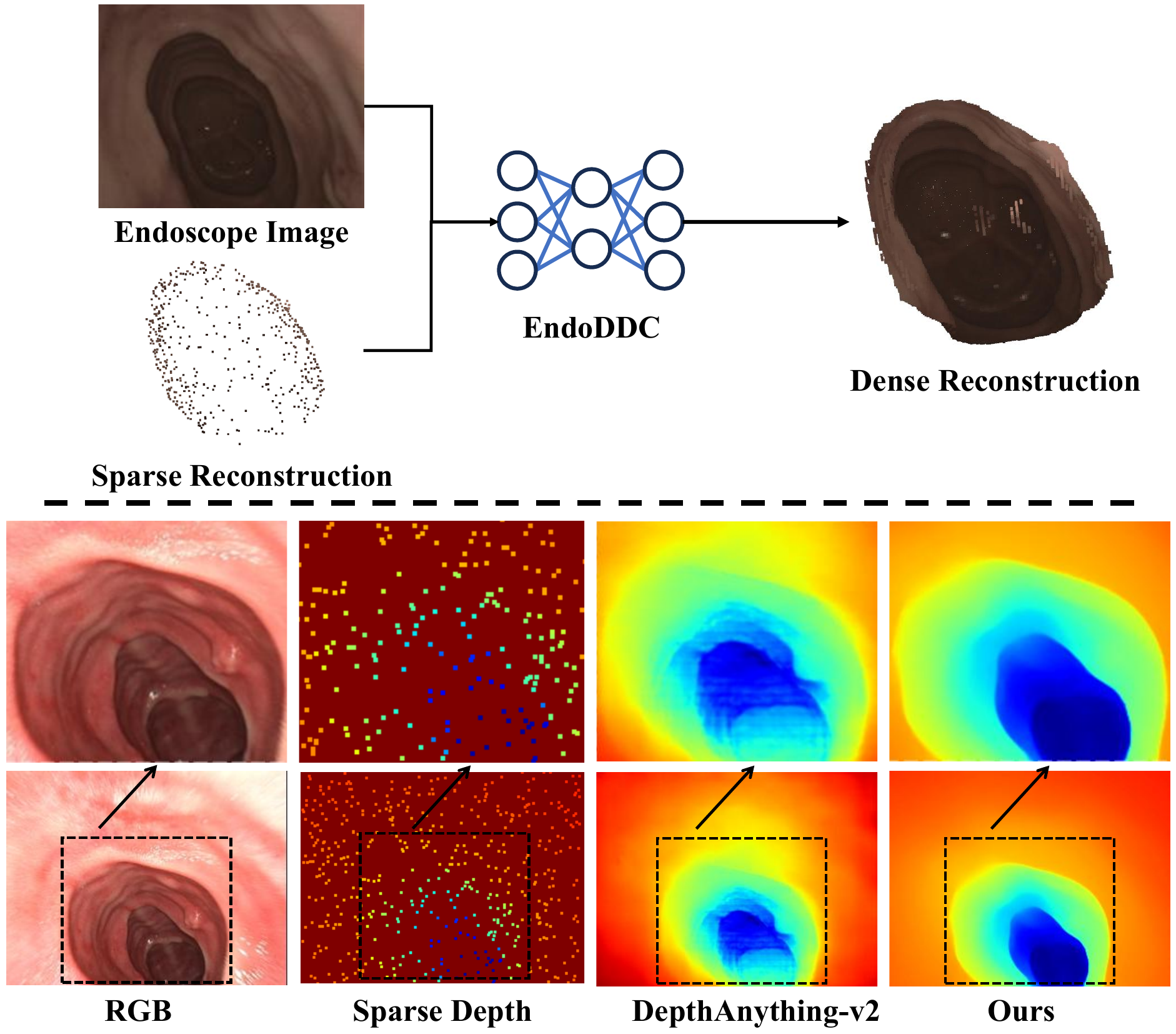}
    \caption[Comparison of SOTA self-supervised regression method (DepthAnything-v2) and our method]{Compared with fine-tuning the SOTA foundational model, our method generates more robust and accurate depth from sparse depth and RGB input, achieving superior sparse to dense reconstruction.}
    \label{fig:intro} 
\end{figure}

Self-supervised regression methods, independent of datasets, provide a promising alternative \cite{ Intro_4.2_Self_Supervised_lightweigh}. However, monocular depth estimation using single RGB images suffers from scale ambiguity and struggles to produce geometrically accurate results, particularly in weakly textured endoscopic environments. Self-supervised methods using video sequences depend on significant camera motion to establish geometric constraints, making them unsuitable for endoscopic scenarios, where minimal camera movement complicates depth and pose estimation \cite{Intro_4.4_LightDepth}.


While self-supervised depth estimation methods suffer from scale ambiguity issues and fail to generate precise depth maps, depth completion with accurate sensor depth becomes a more viable alternative. By leveraging accurate sparse depth points as a geometric prior, depth completion methods address challenges inherent in fine-tuning methods and self-supervised learning, generating dense depth maps by combining sparse depth data with RGB images \cite{BG_2_1}. While depth completion has seen success in fields such as autonomous driving \cite{ognidc}, due to the weak texture and multiple light reflections in endoscopic environments, its application in endoscopic robotic navigation remains unexplored. Additionally, the sparse depth values obtained via depth sensors are difficult to utilize for model fine-tuning.

To tackle these challenges, we introduce EndoDDC, a novel depth completion method tailored to the navigation of endoscopic surgical robots. Our approach integrates depth gradient information and introduces a diffusion model to optimize depth maps, addressing the issues of weak texture and light reflection in endoscopic environments. The method consists of a feature extraction network, a depth fusion module for iterative processing of depth values and gradients, a conditional depth diffusion model for refining the predicted depth map, and an upsampling module for producing high-resolution results. As shown in Fig. \ref{fig:intro}, our method outperforms fine-tuning the current SOTA model, DepthAnything-V2 \cite{depth_anything_v2}, leveraging sparse depth information to compensate for missing gastrointestinal features, significantly improving the geometric and textural details of the intestinal wall.

Our key contributions include the following:
\begin{itemize}
    \item  We propose EndoDDC, a depth completion pipeline that leverages sparse depth and RGB input for sparse to dense reconstruction, addressing challenges posed by fine-tuning and self-supervised learning methods.
    
    \item We propose a multi-scale feature extraction and depth gradient fusion module to provide geometry guidance for sparse to dense endoscopic reconstruction.
    
    \item We propose a depth diffusion strategy with the depth gradient conditioned diffusion model for the endoscopic depth completion task, iteratively optimizing the depth values with coarse depth and depth gradient features.
    

    \item We conduct exhaustive experiments on two public surgical datasets, showing that our model outperforms current SOTA methods by providing more robust sparse to dense reconstruction for endoscopic robotic navigation.
\end{itemize}
\section{ RELATED WORK}

\subsection{Depth Estimation in Endoscopic Surgery}
Recent advances in endoscopic surgery have leveraged the powerful prior knowledge of foundational models~\cite{depth_anything_v2, dust3r_cvpr24, mast3r_eccv24,wang2025vggt} by fine-tuning them on surgical data, enabling precise three-dimensional perception in complex surgical environments. However, the success of this approach heavily depends on the availability of domain-specific datasets with accurate ground truth depth. Due to restrictions related to safety, privacy, and professional regulations, obtaining precise large-scale endoscopic surgery videos with corresponding depth of ground truth remains a significant challenge \cite{Intro_1.4_Early_research,BG_1_1}. As a result, deep learning-based methods for endoscopic depth prediction are shifting toward self-supervised regression.

Simultaneous localization and mapping with multi-view learning, a fundamental paradigm in this domain, is based on the Structure from Motion (SfM). Existing works~\cite{recasens2021endo, rodriguez2024nr, wang2024endogslam, huang2025advancing} utilized representation such as ORB features~\cite{ORBSLAM3_TRO} and 3D Gaussians~\cite{kerbl3Dgaussians}, paving the way in learning-based endoscopic reconstruction. However, the limited range of motion in endoscopic cameras, confined within a cavity, often results in minimal pose changes. This makes the simultaneous estimation of depth and pose an ill-posed problem, diminishing the model’s accuracy and robustness \cite{EndoDAC}. To eliminate reliance on camera motion, self-supervised single-view methods, such as LightDepth \cite{Intro_4.4_LightDepth}, use unpaired single images for training, enabling robust depth estimation even in scenes with minimal or no camera movement. Nevertheless, since depth is inferred solely from single-frame data, the scale of depth values remains unknown. Moreover, variations in scene brightness can cause changes in the scale of each frame, making raw outputs unsuitable for clinical applications that require precise measurements. Therefore, additional mechanisms are necessary to restore the absolute scale \cite{Intro_4.3_Depth_Estimation_Endoscopy}.

\subsection{Depth Completion}

In conventional environments, early depth completion methods directly predicted depth values from images and sparse depth maps \cite{BG_2_1,BG_2_2}. These feedforward approaches struggle with inferring long-range depth due to their limited receptive fields. To improve output quality, post-processing optimization methods based on the spatial propagation network (SPN) mechanism \cite{BG_2_3_SPN} have been continuously proposed and refined. Specifically, CSPN \cite{BG_2_4_CSP} and its successor, CSPN++ \cite{BG_2_5_CSPN++}, introduced convolutions with fixed and adaptive kernels, respectively, improving efficiency. NLSPN \cite{BG_2_6_NLSPN} utilizes deformable convolutions to predict non-local neighborhoods, while DySPN \cite{BG_2_7_DySPN} predicts distinct propagation weights at each iteration, achieving superior performance with fewer iterations and smaller neighborhoods. Earlier architectures primarily processed two-dimensional features, but some methods, such as GuideFormer \cite{BG_2_8_GuideFormer} and CompletionFormer \cite{completionformer}, incorporate camera intrinsic parameters to handle three-dimensional information. Building on these advances, OGNI-DC \cite{ognidc} achieves more robust generalization by iteratively optimizing the depth gradient field.

However, most current methods focus on indoor and outdoor human activity scenarios, with limited efficient depth completion solutions for endoscopic surgery. Liu et al~\cite{ICRA_DC}. proposed a depth completion method for intestinal endoscopy based on multi-scale confidence and self-attention mechanisms, but this method struggles with depth completion in regions with weak textures. To overcome these limitations, we propose a depth completion method based on depth gradients and diffusion optimization. With the physical nature of the diffusion model, our approach ensures precise depth reconstruction across various endoscopic datasets, advancing the dense 3D reconstruction techniques in surgical robotics.
\begin{figure*}
    \centering
    \includegraphics[width=\textwidth]{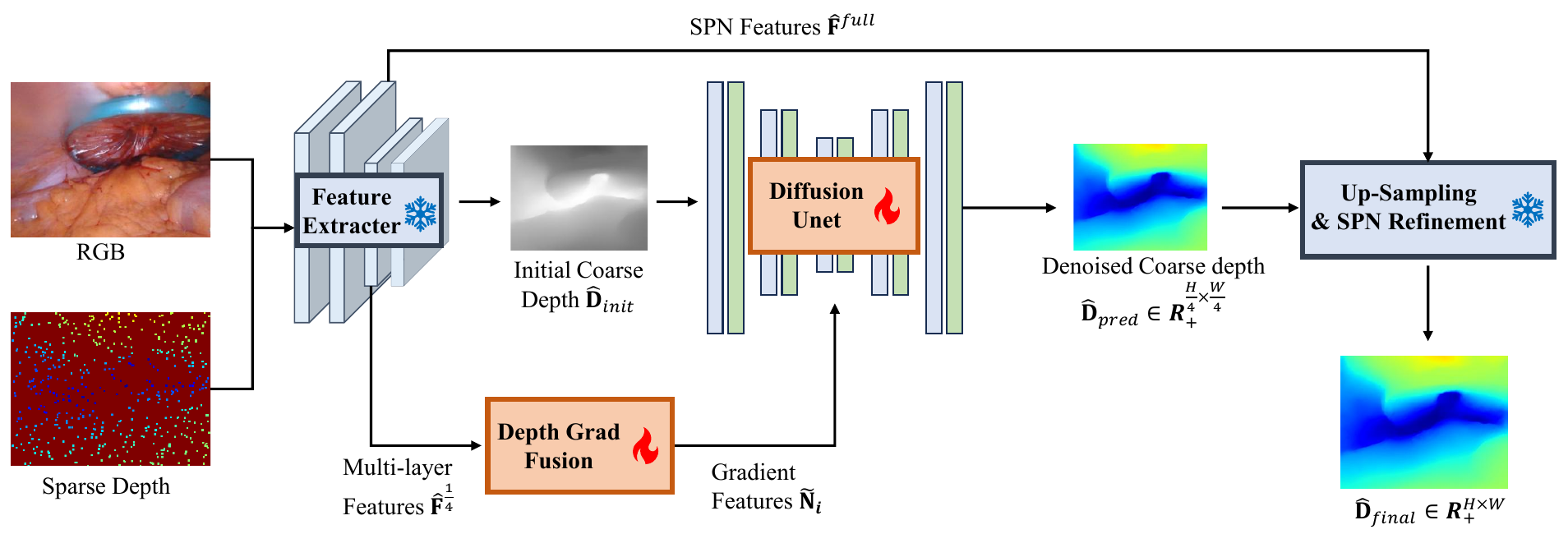}
    \caption{\textbf{Overview of EndoDDC}: After feature extraction from the RGB image and sparse depth map, the Depth Grad Fusion module iteratively updates the state hidden network based on depth and gradient features. This output is then fed into the Depth Diffusion guidance model to conditionally optimize the initial depth. Finally, the optimized coarse depth map undergoes up-sampling and SPN  refinement module to produce the final depth map.}
    \label{fig:model} 
\end{figure*}

\section{Methodology}


EndoDDC takes an RGB image $\mathbf{I}\in\mathbb{R}^{3\times H \times W}$ and its corresponding sparse depth map $\mathbf{S}\in \mathbb{R}_{+}^{H \times W}$ as input, and ultimately outputs a dense depth map. The overall pipeline of our proposed framework is illustrated in Fig. \ref{fig:model}.

\subsection{Preliminary}
Our depth diffusion module builds upon the Denoising Diffusion Implicit Model (DDIM) \cite{DDIM}, a generative method that produces high-fidelity results through a deterministic denoising process. The DDIM framework consists of a forward diffusion process and a reverse denoising process.

\textbf{Forward Diffusion.} The forward process gradually adds Gaussian noise to an initial input $\mathbf{x}_{0}$ over T discrete timesteps. Following a Markov chain at timestep t: $q\left(x_{1: T} \mid x_{0}\right)=\prod_{t=1}^{T} q\left(x_{t} \mid x_{t-1}\right)$,  we have:

\begin{equation}
    q\left(x_{t} \mid x_{t-1}\right)=\mathcal{N}\left(x_{t} ; \sqrt{1-\beta_{t}} x_{t-1}, \beta_{t} I\right)
\end{equation}
where $x_{t}$ indicates the noise data at time-step $t \in[0, T]$, $\beta_{t}$ is the variance and $q\left(x_{1: T} \mid x_{0}\right)$ denotes the probability distribution of $\mathbf{x}_{T}$ obtained after T-step. Based on the above equation, the relationship between $\mathbf{x}_{T}$ and  $\mathbf{x}_{0}$ is:
\begin{equation}
    \mathbf{x}_{t}=\sqrt{\bar{\alpha}_{t}} \mathbf{x}_{0}+\sqrt{1-\bar{\alpha}_{t}} \boldsymbol{\epsilon}_{t}
    \label{2}
\end{equation}
where $\alpha_{t}=1-\beta_{t}$, $\bar{\alpha}_{t}=\prod_{i=0}^{t} \alpha_{i}$ and $\boldsymbol{\epsilon}_{t} \sim \mathcal{N}(\mathbf{0}, \mathbf{I})$.

\textbf{Reverse Denoising.} The reverse process recovers the original data $\mathbf{x}_{0}$ from a noisy input ${\mathbf{x}}_{T} \sim \mathcal{N}(\mathbf{0}, \mathbf{I})$. At each step t, a network $\boldsymbol{\epsilon}_{\theta}$ first predicts the noise component $\boldsymbol{\epsilon}_{t}$ from the noisy input $\mathbf{x}_{T}$, then estimate the initial data as:
\begin{equation}
    \hat{{x}}_{0}\left(x_{t}, t\right)=\frac{x_{t}-\sqrt{1-\bar{\alpha}_{t}} \epsilon_{\theta}}{\sqrt{\bar{\alpha}_{t}}}.
\end{equation}
By the definition of DDIM, we have the denoised data:
\begin{equation}
    \label{5}
    x_{t-1}=\sqrt{\bar{\alpha}_{t-1}} \hat{x}_{0}\left(x_{t}, t\right)+\sqrt{1-\bar{\alpha}_{t-1}} \epsilon_{\theta}({{x}}_{t}, t)
\end{equation}


This deterministic process is equivalent to solving a probability flow Ordinary Differential Equation (ODE) \cite{DDIM_ODE_1,DDIM_ODE_2}. Its iterative correction mechanism effectively rectifies errors caused by endoscopic challenges, like textureless surfaces or specular reflections, ensuring a globally coherent and physically plausible final depth map.

\subsection{Feature Extraction and Depth Grad Fusion}
In feature extraction, we first encode the input image $\mathbf{I}$ and the sparse depth map $\mathbf{S}$ through a series of convolutional layers. After concatenation, we use the pre-trained PVT~\cite{wang2021pyramidvisiontransformerversatile} to extract both global and local information, generating feature maps at multiple scales. After encoding, the network enters the decoding stage, where it gradually restores spatial resolution through upsampling and skip connections. The network ultimately outputs the initial coarse depth prediction and features at two resolution scales: 1/4 resolution for intermediate depth predictions and full resolution for final depth enhancement with SPN:
    \begin{equation}
	\hat{\mathrm{D}}_{\text {init }}, \hat{\mathrm{F}}^{\text {full }}, \hat{\mathrm{F}}^{\frac{1}{4}}=\text {Backbone}(\mathbf{I},\mathbf{S})
    \end{equation}
Following ~\cite{ognidc}, we use CompletionFormer \cite{zhang2023completionformer} as the backbone of our feature extraction and initialize $\hat{\mathbf{G}_{0}}$ with $0$.

\begin{figure*}
    \centering
    \includegraphics[width=\textwidth]{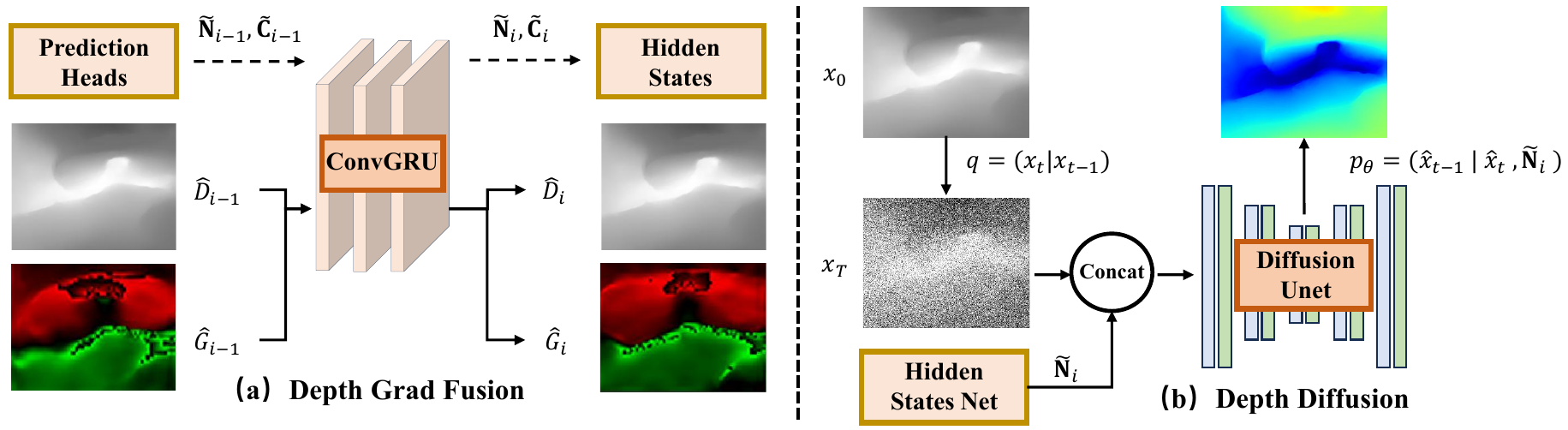}
    \caption{\textbf{Details of the proposed model.} (a) Details of the Depth Grad Fusion module; (b) The architectural design of the Depth Diffusion module.}
    \label{fig:details of model} 
\end{figure*}

Fig. \ref{fig:details of model}(a) shows that the Depth Grad Fusion module iteratively processes depth maps and gradient maps using convolutional gated recurrent units (ConvGRU)\cite{teed2020raftrecurrentallpairsfield} to obtain more precise feature networks containing depth and gradient information:
    \begin{equation}
	\Delta \hat{\mathbf{D}_{i}},\Delta \hat{\mathbf{G}_{i}}, \tilde{\mathbf{N}}_{i}=\operatorname{ConvGRU}\left(\tilde{\mathbf{N}}_{i-1}, \tilde{\mathbf{C}}_{i-1}, \hat{\mathbf{D}}_{i-1}, \hat{\mathbf{G}}_{i-1}\right)
    \end{equation}
For each step, this model takes hidden state net $\tilde{\mathbf{N}}_{i-1}$, context input $\tilde{\mathbf{C}}_{i-1}$, predicted depth gradient $\hat{\mathbf{G}}_{i-1}$, and predicted depth $\hat{\mathbf{D}}_{i-1}$ as input to generate the new hidden, variation of the context and depth gradient. The hidden state net $\tilde{\mathbf{N}}_{0}$ and init context input $\tilde{\mathbf{C}}_{0}$ are obtained from backbone features $\hat{\mathrm{F}}^{\frac{1}{4}}$. With the variation, we have the current depth $\hat{\mathbf{D}_{i}}=\Delta \hat{\mathbf{D}_{i}}+\hat{\mathbf{D}}_{i-1}$ and current depth gradient $\hat{\mathbf{G}_{i}}=\Delta \hat{\mathbf{G}}_{i}+\hat{\mathbf{G}}_{i-1}$. The final output of the hidden state net $\tilde{\mathbf{N}}_{i}$ is utilized as the depth diffusion condition, and the depth gradient $\hat{\mathbf{G}}_{i}$ serves as an intermediate for the depth gradient fusion optimization.

\subsection{Depth Diffusion}
The iterative denoising process of DDIM enforces a smooth path from noise to data, ensuring the global coherence of the final depth map. However, while this mechanism is effective in maintaining overall plausibility, it lacks contextual initialization and explicit guidance to resolve the local ambiguities inherent in endoscopic imagery. To constrain the solution space and guide the generation process toward the correct geometry, we first utilize the initial coarse depth $\hat{\mathbf{D}}_{init}$ as the initialization ${x}_{t}$ for the reverse denoising. 

Then we introduce the explicit geometric diffusion condition by utilizing depth gradient features, which encode crucial information about local surface structure and orientation. As shown in Fig. \ref{fig:details of model}(b), we denote the hidden states net containing depth and gradient information, extracted by a preceding network, as $\tilde{\mathbf{N}}_i\in\mathbb{R}^{C \times \frac{H}{4} \times \frac{W}{4}}$. And we distill this high-dimensional feature map into a single-channel guidance map $\tilde{\mathbf{N}}^\prime_i\in\mathbb{R}^{1 \times \frac{H}{4} \times \frac{W}{4}}$  using a $1\times1$ convolution. This projection preserves the essential geometric cues while matching the dimensions of the depth map.

Then we incorporate this guidance by concatenating $\tilde{\mathbf{N}}_{i}^\prime$ with the noisy depth map ${x}_{t}$ along the channel axis. The combined tensor serves as the input to our noise prediction network. This modification explicitly informs the network about the underlying geometry at each denoising step, enabling it to make more accurate noise predictions. The conditional noise prediction is thus formulated as follows:

\begin{equation}
    \boldsymbol{\epsilon}_{\theta} = \boldsymbol{\epsilon}_{\theta}(concat({{x}}_{t}, \tilde{\mathbf{N}}_i^\prime), t)
\end{equation}

By training the network with this objective, our depth diffusion model learns to leverage geometric priors to effectively optimize the depth map. And the process of obtaining depth maps through iterative denoising, expressed in equation \ref{5}, can be rewritten as:
\begin{equation}
    \begin{split}
        x_{t-1} = {} & \sqrt{\bar{\alpha}_{t-1}} \hat{x}_{0}(x_{t}, t) + \sqrt{1-\bar{\alpha}_{t-1}} \\
        & \boldsymbol{\epsilon}_{\theta}(concat({{x}}_{t}, \tilde{\mathbf{N}}_i^\prime), t)
    \end{split}
\end{equation}

In the training phase, since $\bar{\alpha}_{t}$, $t$, $\tilde{\mathbf{N}}$ and ${{x}}_{t}$ are known, we only need to train the network $\boldsymbol{\epsilon}_{\theta}$ to accurately predict the noise $\boldsymbol{\epsilon}_{t}$ to get the corrected depth $\hat{\mathbf{D}}_{pred}\in\mathbb{R}^{\frac{H}{4} \times \frac{W}{4}}_{+}$.

\subsection{Depth Enhancement and Loss Function}
\textbf{Depth Up-Sampling and SPN Refinement.} We use the convex combination method \cite{teed2020raftrecurrentallpairsfield} to up-sample the 1/4 resolution depth prediction map $\hat{\mathbf{D}}_{pred}\in\mathbb{R}^{\frac{H}{4} \times \frac{W}{4}}_{+}$ to obtain the full resolution depth map $\hat{\mathbf{D}}_{pred}\in\mathbb{R}^{H \times W}_{+}$, and then use a pre-trained spatial propagation network (SPN) \cite{SPN} to augment the up-sampled depth map to obtain the final depth prediction map $\hat{\mathbf{D}}_{final}$:
    \begin{equation}    \hat{\mathbf{D}}_{final}=\operatorname{SPN}\left(\hat{\mathbf{D}}_{pred}, \hat{\mathbf{F}}^{\text {full }}\right)
\end{equation}

\textbf{Depth Loss.} During the training process, we combine $L_{1}$ and $L_{2}$ loss to supervise the final depth maps $\hat{\mathbf{D}}_{pred}$ and the up-sampled depth maps $\hat{\mathbf{D}}_{final}$. Given the ground-truth depth map D, the loss function of depth $\mathcal{L}_{\mathrm{D}}$ is: 
\begin{equation}
    \begin{split}
        \mathcal{L}_{\mathbf{D}} ={}& \left\|\hat{\mathbf{D}}_{pred}-\mathbf{D}\right\|_{2}^{2}+\left\|\hat{\mathbf{D}}_{pred}-\mathbf{D}\right\|_{1} \\
        & +\left\|\hat{\mathbf{D}}_{final}-\mathbf{D}\right\|_{2}^{2}+\left\|\hat{\mathbf{D}}_{final}-\mathbf{D}\right\|_{1}
    \end{split}
\end{equation}

\textbf{Gradient Loss.} In down-sampling, we compute the corresponding gradient $G$ based on the provided ground-truth depth map $D$ and supervise the depth gradients with $L_{1}$:
\begin{equation}
    \mathcal{L}_{\mathbf{G}}=\gamma^{I-i}\left\|\hat{\mathbf{G}}_{i}-\mathbf{G}\right\|_{1}
\end{equation}
where I is the total number of iterations for convgru, i represents the current iteration, and loss decay $\gamma$ is 0.9.

\textbf{Depth Diffusion Loss.} In the training of the depth diffusion model, we optimize $\boldsymbol{\epsilon}_{\theta}(concat({{x}}_{t}, \tilde{\mathbf{N}}_i^\prime), t)$ to promote it close to the Gaussian noise, and this loss is supervised with $L_{2}$ and can be written as:
\begin{equation}
    \mathcal{L}_{diff}=\left\|\boldsymbol{\epsilon}_{t}-{\epsilon}_{\theta}(concat({{x}}_{t}, \tilde{\mathbf{N}}_i^\prime), t)\right\|_{2}
\end{equation}

\section{Experiments}

\begin{figure*}[ht!]
    \centering
    \includegraphics[width=\textwidth]{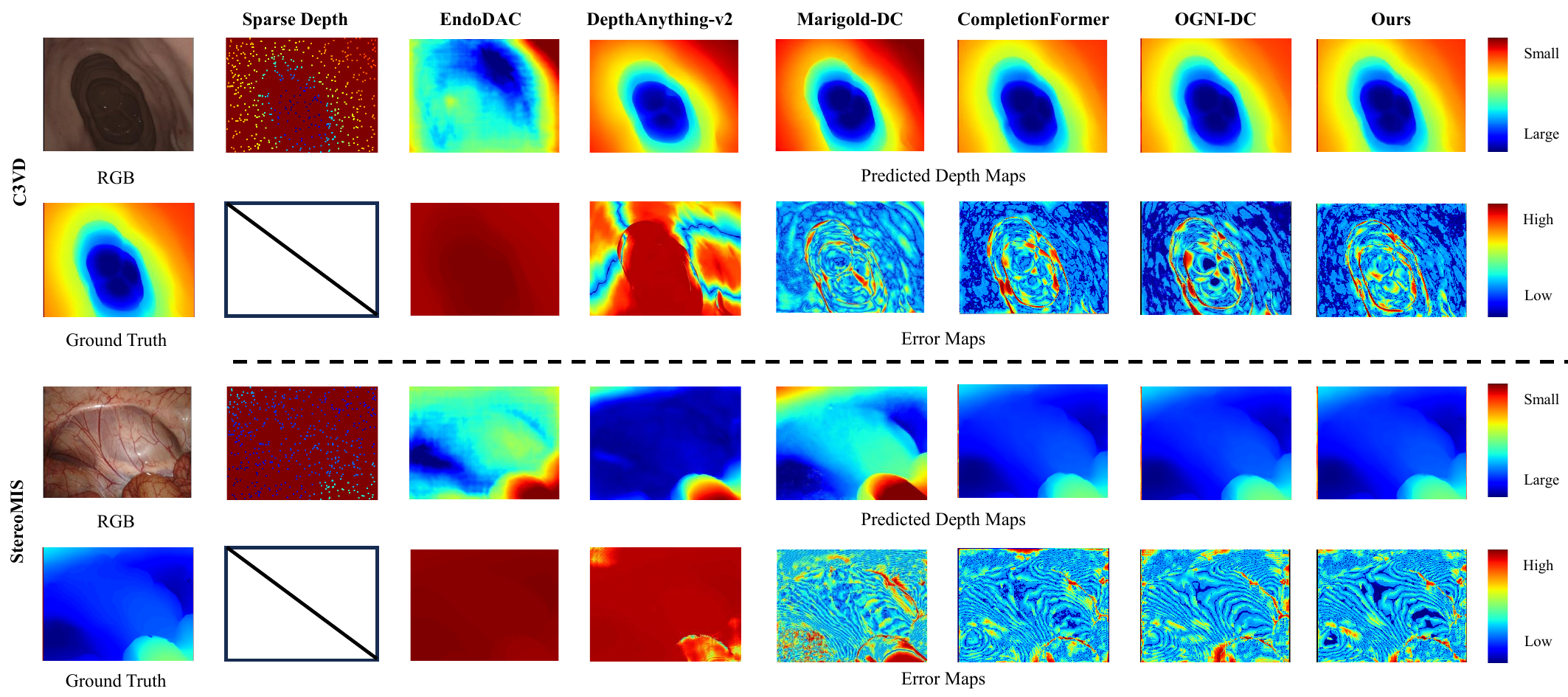}
    \caption[]{\textbf{Qualitative comparison on C3VD and StereoMIS datasets.} We compare EndoDDC with SOTA depth estimate and depth completion methods; our method generates less error in terms of tissue details.}
    \label{fig:result} 
\end{figure*}

\begin{table*}[ht!]
    \centering
    \caption{Quantitative depth comparison with SOTA methods on C3VD~\cite{c3vd} and StereoMIS~\cite{StereoMIS} dataset.}
    \label{tab:c3vd_comparison} 
    \resizebox{\textwidth}{!}{%
        \begin{tabular}{l|cccc|cccc}
            \toprule
            \multicolumn{1}{l|}{\multirow{2}{*}{\centering Models}} & \multicolumn{4}{c|}{C3VD~\cite{c3vd}} & \multicolumn{4}{c}{StereoMIS~\cite{StereoMIS}}\\ \cmidrule{2-9} 
                                    & RMSE (mm) ↓          & MAE (mm) ↓           & REL (mm) ↓           & $\delta$ ↑  & RMSE (mm) ↓          & MAE (mm) ↓           & REL (mm) ↓           & $\delta$ ↑           \\ \midrule
            EndoDAC~\cite{EndoDAC}                 & 9.7476          & 7.5541          & 0.1081          & 0.9162  & 11.8435         & 8.7066          & 0.2084          & 0.6765        \\
            DepthAnything-v2~\cite{depth_anything_v2}        & 5.2202          & 3.6901          & 0.0671          & 0.9892  & 2.2465          & 1.63765         & 0.0277          & 0.8857        \\
            Marigold-DC~\cite{marigolddc}             & 0.8294          & 0.3275          & 0.0106          & 0.9985  & 2.6755          & 1.0395          & 0.0122          & 0.9968        \\
            CompletionFormer~\cite{completionformer}        & 0.6875          & 0.2320          & 0.0071          & 0.9988  & 1.6536          & 0.6261          & 0.0069          & 0.9986        \\
            OGNI-DC~\cite{ognidc}                 & 0.6770          & 0.2283          & 0.0067          & 0.9988  & 1.5857          & 0.6114          & 0.0063          & 0.9986        \\
            Ours                    & \textbf{0.6412} & \textbf{0.2104} & \textbf{0.0060} & \textbf{0.9990} & \textbf{1.4691} & \textbf{0.5515} & \textbf{0.0061} & \textbf{0.9988}\\ \bottomrule
        \end{tabular}%
    }
\end{table*}

\subsection{Datasets and Evaluation Metrics}
\textbf{C3VD} (Colonoscopy 3D Video Dataset) \cite{c3vd} is a 3D colonoscopy video dataset acquired using high-definition clinical colonoscopes and high-fidelity colon models. The dataset comprises 10,015 annotated images and 22 registered video sequences, along with corresponding ground truth for depth, surface normals, optical flow, occlusion, six-degree-of-freedom (6-DoF) poses, coverage maps, and 3D models. In our experiments, we trained and tested C3VD using the dataset splits proposed in LightDepth \cite{c3vdsplit}.

\textbf{StereoMIS} \cite{StereoMIS} is a dataset designed for SLAM applications in endoscopic surgery. It was captured using the da Vinci Xi surgical robot to record stereo video streams and forward camera motions across three live porcine subjects. It encompasses diverse anatomical structures and challenging scenarios involving significant tissue deformation. We extracted images from the video frames at a frame ratio of 10:1 for its 11 scene videos. We used the images extracted from video frames P1 through P2$\_$7 as the training set and the images extracted from P2$\_$8 as the test set. In addition, since the dataset does not provide the ground truth of the paired videos, we refer to Deform3DGS \cite{get_StereoMIS_Depth} for depth estimation of the videos to obtain ground true depths of the images.

\textbf{Evaluation Metrics.} We evaluate under 4 standard metrics \cite{metric1,metric2,DPsurvey}: root mean squared error RMSE, mean absolute error MAE, mean absolute relative error REL in millimeters (mm), and $\delta$ accuracy, which measures the percentage of predicted pixels $\hat{d}$ that satisfy $\max(\frac{\hat{d}}{d}, \frac{d}{\hat{d}}) < 1.25^k$, where $k=1$. This metric provides a robust measure of the proportion of depth estimates within a clinically acceptable error margin.

\subsection{Implementation Details}
We implement the proposed method with PyTorch on 2 NVIDIA RTX 4090 GPUs for 36 epochs with a batch size of 6. The dimensions of the images used in the model are 256$\times$320. We employ the AdamW optimizer for optimization with the initial learning rate set to 0.001. For the depth gradient fusion, we use $i=5$ steps. For the depth diffusion model, we use the U-Net  \cite{Unet}  architecture as the noise estimation network, and the time step $T$ and sampling step $S$ for the forward diffusion and backward denoising processes were set to 1000 and 20, respectively. For fair comparison, we create sparse depth maps following~\cite{ICRA_DC}.

\begin{figure*}[ht!]
    \centering
    \includegraphics[width=\textwidth]{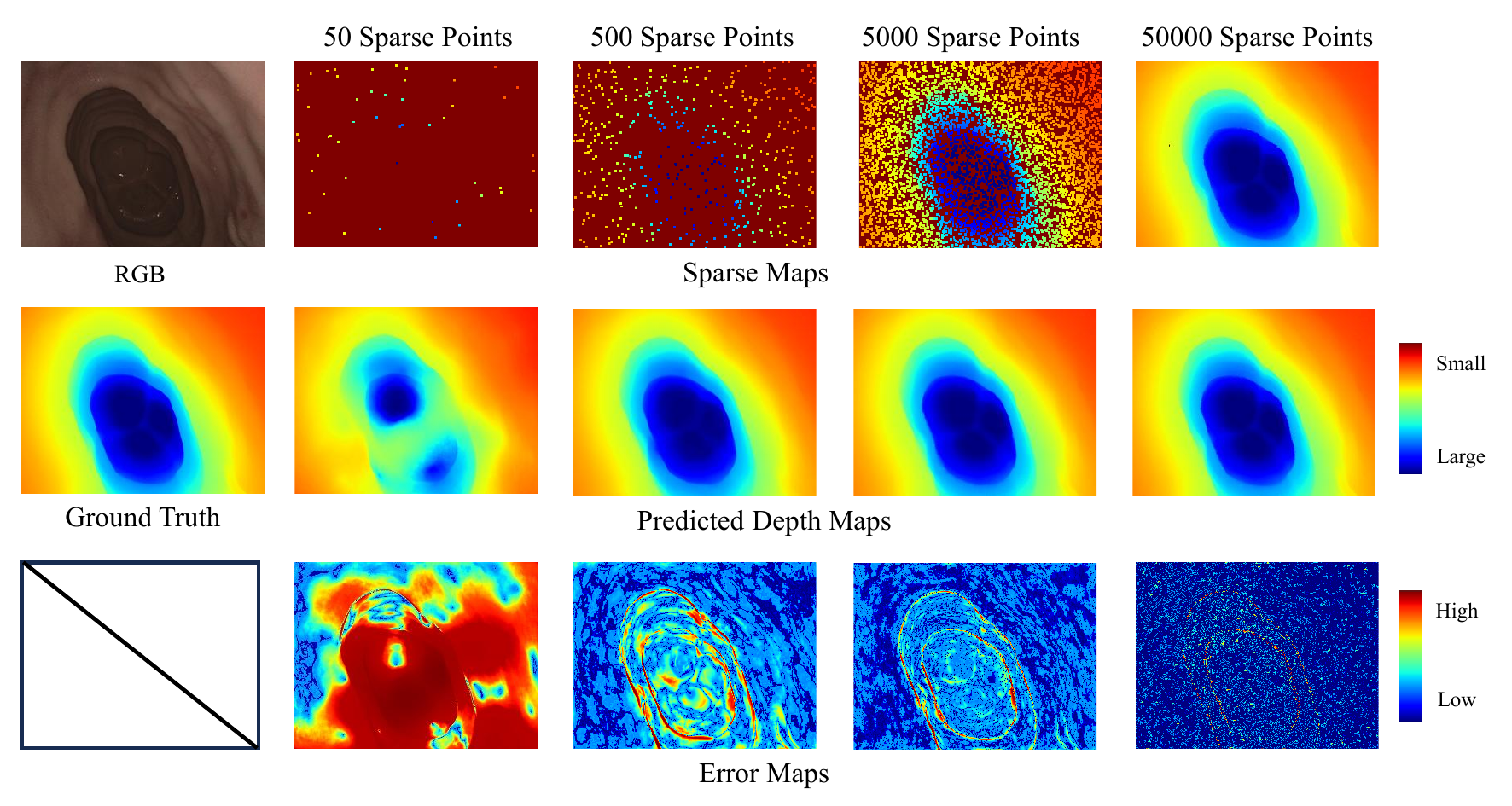}
    \caption[\textbf{Results of EndoDDC at different levels of sparsity}]{\textbf{Results of EndoDDC at different levels of sparsity}. Our method generates robust dense depth across different levels of sparsity.}
    \label{fig:different_sparsities} 
\end{figure*}
\begin{table*}[ht!]
    \centering
    \caption{Quantitative Comparison Results of Depth Completion Models at Different Sparsity Levels on C3VD~\cite{c3vd} dataset.}
    \label{tab:sparse} 
    \resizebox{\textwidth}{!}{%
        \begin{tabular}{l|cc|cc|cc|cc}
            \toprule
            \multicolumn{1}{l|}{\multirow{2}{*}{\centering Methods}} & \multicolumn{2}{c|}{50} & \multicolumn{2}{c|}{500}& \multicolumn{2}{c|}{5000} & \multicolumn{2}{c}{50000}\\ \cmidrule{2-9} 
                                    & RMSE (mm) ↓          & REL (mm) ↓           & RMSE (mm) ↓           &  REL (mm) ↓ & RMSE (mm) ↓          & REL (mm) ↓           & RMSE (mm) ↓           & REL (mm) ↓         \\ \midrule
            CompletionFormer~\cite{completionformer}                & 4.9533          & 0.0717          & 0.6875          & 0.0071  & 0.5848         & 0.0064          & 5.7541          & 0.1629        \\
            Marigold-DC~\cite{marigolddc}        & \textbf{1.8955}          & \textbf{0.0302}          & 0.8294          & 0.0106  & 0.5596          & 0.0091         & 0.4569          & 0.0088        \\
            OGNI-DC~\cite{ognidc}             & 3.6632          & 0.0767          & 0.6770          & 0.0067  & 0.3757          & 0.0029          & 0.1403          & 0.0007        \\
            Ours                    & 3.6772 & 0.0703 & \textbf{0.6591} & \textbf{0.0064} & \textbf{0.3442} & \textbf{0.0025} & \textbf{0.1177} & \textbf{0.0004}\\ \bottomrule
        \end{tabular}%
    }
\end{table*}


\begin{table*}[ht!]
    \centering
    \caption{Ablation Study on C3VD~\cite{c3vd} and StereoMIS~\cite{StereoMIS} dataset. "Baseline" indicates using ~\cite{completionformer, teed2020raftrecurrentallpairsfield} without the Depth Grad Fusion and Depth Diffusion. "w/o" denotes "without". DDI indicates Depth-Depth Iteration~\cite{ognidc}.}
    \label{tab:ablation} 
    \resizebox{\textwidth}{!}{%
        \begin{tabular}{l|cccc|cccc}
            \toprule
            \multicolumn{1}{l|}{\multirow{2}{*}{\centering Methods}} & \multicolumn{4}{c|}{C3VD~\cite{c3vd}} & \multicolumn{4}{c}{StereoMIS~\cite{StereoMIS}}\\ \cmidrule{2-9} 
                                    & RMSE (mm) ↓          & MAE (mm) ↓           & REL (mm) ↓           & $\delta$ ↑  & RMSE (mm) ↓          & MAE (mm) ↓           & REL (mm) ↓           & $\delta$ ↑           \\ \midrule
            Baseline             & 0.6883          & 0.2329          & 0.0069          & 0.9987  & 1.6159          & 0.6163          & 0.0065          & 0.9985        \\
            Baseline+DDI~\cite{ognidc}        & 0.6770          & 0.2283          & 0.0067          & 0.9988  & 1.5857          & 0.6114          & 0.0063          & 0.9986        \\

            Ours (w/o Depth Grad)        & 0.6669          & 0.2304          & 0.0068          & 0.9988  & 1.5729          & 0.5893          & 0.0064          & 0.9985        \\
            
            Ours (w/o Init Depth)        & 0.6453          & 0.2261          & 0.0063          & 0.9983  & 1.5604          & 0.5675          & 0.0062          & 0.9985        \\
            Ours Full                   & \textbf{0.6412} & \textbf{0.2104} & \textbf{0.0060} & \textbf{0.9990} & \textbf{1.4691} & \textbf{0.5515} & \textbf{0.0061} & \textbf{0.9988}\\ \bottomrule
        \end{tabular}%
    }
\end{table*}

 \subsection{Results}
The proposed method is compared with six SOTA baselines: DepthAnything-v2 \cite{depth_anything_v2} and EndoDAC \cite{EndoDAC} address the task of depth estimation, focusing on conventional and surgical scenes, respectively. Meanwhile, CompletionFormer \cite{completionformer}, Marigold-DC \cite{marigolddc}, and OGNI-DC \cite{ognidc} are leading methods representative of the depth completion task within conventional environments. Results are shown in Table \ref{tab:c3vd_comparison}, and Fig.\ref{fig:result}. From the comparisons in Table \ref{tab:c3vd_comparison}, our method outperforms all metrics on both datasets. For the accuracy metric $\delta$, our method improves compared to all other methods, especially compared to EndoDAC in the StereoMIS dataset, where our method improves the accuracy by $25.55\%$. For the error metrics, compared to the fine-tuned DepthAnything-v2 and self-supervised learning depth estimate models EndoDAC, our method reduces RMSE, MAE, and REL metrics substantially on both datasets, even compared to the best depth completion model OGNI-DC, RMSE, MAE, and REL on C3VD are reduced respectively by $5.28\%$, $7.84\%$, and $10.44\%$, while on StereomMIS, they decreased by $7.35\%$, $9.79\%$, and $3.17\%$, respectively.

In Fig.\ref{fig:result}, we observe that the fine-tuning DepthAnything-v2 and EndoDAC perform poorly in terms of millimeter-level depth estimation when compared to other depth completion methods. This is because, compared to the depth completion model, this class of methods lacks the constraints of a sparse depth map during training and can only estimate the depth value through the relationship between image frames, resulting in a weaker depth prediction capability compared to depth completion models. Additionally, when compared to the other three depth completion models, Marigold-DC, CompletionFormer, and OGNI-DC, our method demonstrates superior depth prediction capabilities, especially at image edges and regions with significant depth variations. In the error maps, we can find that our method exhibits smaller estimation errors at the edges of the image and in regions with large depth gradient variations. In general, our approach outperforms the other SOTA methods.

\subsection{Robustness to Different Sparse Maps}
For downstream applications in surgical robotics, a model's robustness to varying levels of sparsity is of critical importance \cite{min2025innovating}. However, retraining separate models for different sparsity levels is impractical for real-world applications, as the degree of depth sparsity cannot be known a priori in an actual surgical environment. For example, in robot-assisted minimally invasive surgery, the acquired sparse depth observations can fluctuate drastically due to occlusions by surgical instruments, specular reflections from tissue surfaces, or changing endoscopic viewpoints. Therefore, we aim to build a single model that performs well across all sparsity levels.

Following~\cite{ICRA_DC}, we uniformly trained the models on the C3VD~\cite{c3vd} under 500 sparse points and evaluated their performance at 50, 500, 5000, and 50000 sparse points, comparing them with multiple baseline models. Fig. \ref{fig:different_sparsities} shows our model's test results across different sparsity levels. As Table~\ref {tab:sparse} indicates, except at 50 sparse points, our model consistently outperforms other baseline models under identical sparsity conditions. This is because EndoDDC, as an end-to-end model specifically designed for endoscopic robots, has an architecture optimized for deeply integrating RGB features with sparse geometric input information. When sparse inputs are drastically reduced to 50 points, the geometric constraints provided become extremely weak. In contrast, the baseline model of Marigold-DC -- Stable Diffusion~\cite{Stable_Diffusion}, trained on billions of images, leverages a powerful visual prior pre-trained on massive datasets to generate relatively well-structured depth maps even under such severely underconstrained conditions. However, the nature of the depth completion task relies not on generation driven by visual prior, but on high-precision reconstruction based on dense geometric constraints. Consequently, as the number of sparse points increases, EndoDDC demonstrates superior performance compared to other models.

\subsection{Ablation Study}
In the ablation study, we compare our method with the baseline~\cite{completionformer, teed2020raftrecurrentallpairsfield} and the DDI~\cite{ognidc} method for iterative update. For fair comparison, we report the results of the baseline without the Depth Diffusion and with DD. To enable fair comparison, we fine-tune the baseline method on the surgical datasets~\cite{c3vd, StereoMIS}. Specifically, we initialized the baseline models with weights pre-trained on natural image datasets and subsequently trained them on C3VD and StereoMIS using the same optimizer and learning rate as EndoDDC. This process ensures that the performance gap is not due to domain shift between natural and surgical environments, but rather stems from the architectural advantages of our proposed diffusion-based refinement. We also compare our full method with the version using depth diffusion without the Depth Grad Fusion feature as diffusion condition, and the version of depth diffusion with depth gradient condition but without the initial depth. The former performs diffusion without guidance of depth and gradient features, while the latter does not incorporate initial depth information during denoising, instead initiating the diffusion process directly from random noise. The results are presented in Table~\ref{tab:ablation}. Compared with others, our method shows dramatic improvement across all metrics. This result shows that our proposed Depth Grad Fusion module (for diffusion conditioning) and Depth Diffusion module with initial depth significantly optimize the depth map, leading to state-of-the-art performance by effectively modeling the data distribution and reducing residual errors.

\section{Conclusions}
In this paper, we propose EndoDDC, an advanced endoscopy depth completion method that significantly improves depth estimation accuracy by introducing sparse depth information and using depth and depth gradient features for conditioning diffusion optimization. Experimental results on two publicly available endoscopy datasets show that the proposed method outperforms the state-of-the-art in all fine-tuning methods, self-supervised depth estimation, and depth completion tasks, particularly in handling complex endoscopic environments with weak textures and challenging lighting conditions. By providing an accurate and robust depth completion method, this work has the potential to significantly enhance the autonomous navigation capabilities of surgical robots, improving procedural safety and efficiency.



\bibliographystyle{IEEEtran}
\bibliography{references}

\end{document}